# Comparison of different Artificial Neural Networks for Bitcoin price forecasting

Silas Baumann
Department of Electrical Engineering and Computer Science
KIT
Karlsruhe, Germany

Karl A. Busch
Department of Electrical Engineering and Computer Science
KIT
Karlsruhe, Germany

Hamza A. A. Gardi
Department of Electrical Engineering and Computer Science
KIT
Karlsruhe, Germany

*Abstract*—**This study investigates the impact of varying sequence lengths on the accuracy of predicting cryptocurrency returns using Artificial Neural Networks (ANNs). Utilizing the Mean Absolute Error (MAE) as a threshold criterion, we aim to enhance prediction accuracy by excluding returns that are smaller than this threshold, thus mitigating errors associated with minor returns. The subsequent evaluation focuses on the accuracy of predicted returns that exceed this threshold. We compare four sequence lengths—168 hours (7 days), 72 hours (3 days), 24 hours, and 12 hours—each with a return prediction interval of 2 hours. Our findings reveal the influence of sequence length on prediction accuracy and underscore the potential for optimized sequence configurations in financial forecasting models.**

## I. Introduction

Since Bitcoin was introduced in 2008 as a digital peer-to-peer equivalent currency built on blockchain technology [1], it emerged as a financial asset that is nowadays mainly used for investments [2].

The forecasting of time series data, such as the Bitcoin price, is a well-known problem existing in many different domains. Depending on the type of data being predicted, the difficulty of achieving an accurate result varies. For example, the prediction of the next sunrise time is relatively easy, whereas tomorrow's winning lottery numbers cannot be predicted with any accuracy. There are many methods for time series forecasting, ranging from classical mathematical models to approaches using deep neural networks and deep learning [3]. In this paper, the data-driven approach of forecasting by applying different types of Artificial Neural Network (ANN) is used.

We try to predict the future price movement of Bitcoin just by reviewing the past market data and compare the performance of the different ANNs on this task based on their predictions.

The paper addresses the question of whether and how well different types of ANNs can predict the future Bitcoin price returns based on current market data.

Methods and results of previous works are discussed in the following chapter. Then the used methods are explained. The results achieved with the previously presented models are shown and compared to each other. Finally, the work is concluded and a brief overview of what can be accomplished with this kind of models is given.

## II. Related Work

In this chapter, we examine fundamental concepts of time series analysis as well as the use of PyTorch for implementing time series models. We draw insights from two sources: "Forecasting: Principles and Practice" by Rob J. Hyndman and George Athanasopoulos [3], as well as a tutorial on time series analysis from TensorFlow [4].

### A. Time Series Analysis

Time series analysis involves the examination and prediction of data collected over time. Specific patterns and trends in the data are identified to forecast future developments. This form of analysis finds application in various fields such as economics, finance, climatology, and healthcare. Time Series Models and Networks

Time series models, including neural networks, are used for modeling time series data. These models capture complex dependencies and patterns in the data to make accurate predictions.Neural networks such as recurrent neural networks (RNNs) and Long Short-Term Memory (LSTM) networks are particularly suitable for considering temporal dependencies in the data and are therefore widely used in time series analysis.

### B. Stationarity

Stationarity is a fundamental concept in time series analysis, indicating, that a time series exhibits constant statistical properties over time, such as mean, variance, and autocorrelation. This assumption simplifies the modeling and forecasting of time series data. Analyzing stationarity is crucial for developing appropriate models for the data, as many time series models require stationarity as a prerequisite.

*1) Augmented Dickey-Fuller Test:* The Augmented Dickey-Fuller (ADF) test is a widely used statistical test to check the stationarity of a time series. The null hypothesis of the test states, that the time series have a unit root, which means it is not stationary. Rejecting the null hypothesis suggests, that the time series is stationary.

The ADF test works by estimating the following regression:

$$\Delta y_t = \alpha + \beta t + \gamma y_{t-1} + \delta_1 \Delta y_{t-1} + \delta_2 \Delta y_{t-2} + \cdots + \delta_p \Delta y_{t-p} + \epsilon_t \quad (1)$$

where:

- $y_t$ is the time series.
- $\Delta$ is the difference operator.
- $\alpha$ is a constant.
- $\beta t$ represents a time trend.
- $\gamma$ is the coefficient of $y_{t-1}$, crucial for testing stationarity.
- $\delta_i$ are the coefficients of the lagged differences.
- $\epsilon_t$ is the error term.

The critical aspect of the ADF test is the t-statistic for the coefficient $\gamma$. If this statistic is sufficiently negative, the null hypothesis of a unit root can be rejected, indicating that the time series is stationary.

The key formulas of the ADF test can be summarized as follows:

1) **Difference Operator**:

$$\Delta y_t = y_t - y_{t-1} \quad (2)$$

This formula calculates the first difference of the time series, an essential step to obtain stationary data.

2) **ADF Regression**:

$$\Delta y_t = \alpha + \beta t + \gamma y_{t-1} + \delta_1 \Delta y_{t-1} + \delta_2 \Delta y_{t-2} + \cdots + \delta_p \Delta y_{t-p} + \epsilon_t \quad (3)$$

This regression includes all relevant terms to model the dynamics of the time series and test for a unit root.

The ADF test is particularly useful for identifying the presence of trends or non-stationarity in a time series and performing suitable transformations, such as differencing, to achieve stationarity before further analysis or modeling. By analyzing the test statistics, it can be determined whether to reject the null hypothesis and thus consider the time series as stationary.

*C. PyTorch for Time Series Analysis*

PyTorch is a powerful deep learning library known for its dynamic computation graph and flexibility. Compared to other frameworks like Keras or TensorFlow, PyTorch offers several advantages:

- **Dynamic Computation Graph**: PyTorch constructs computation graphs dynamically, which facilitates modeling complex structures and simplifies debugging.
- **Flexibility**: PyTorch provides high flexibility for implementing custom models and layers, which is particularly beneficial in research.
- **Integration with Python**: The seamless integration with Python makes it easier to use PyTorch with other scientific libraries and tools.

In time series analysis, PyTorch offers robust tools for implementing and training time series models, especially neural networks like RNNs and LSTMs, which are specifically designed for modeling temporal dependencies.

## III. PROPOSED METHOD

*A. Data Acquisition*

The BTC/USDT trading pair is selected for this study for several reasons. First, Bitcoin (BTC) is the most widespread and well-known cryptocurrency, which has a high trading volume and significant liquidity. These characteristics make BTC an ideal candidate for quantitative analyses and prediction models, as a larger amount of data and more frequent trading activities can increase the accuracy of the models.

Second, USDT (Tether) is a so-called stablecoin, whose value is pegged to the US dollar. This stability reduces volatility compared to other cryptocurrencies, simplifying the analysis and prediction of price movements. Thus, the BTC/USDT pair combines the advantages of a highly liquid asset with the stability of a stablecoin, leading to more robust and reliable models.

Furthermore, the Binance exchange provides comprehensive historical data for BTC/USDT, allowing access to a rich dataset. This data availability is crucial for the development and training of accurate prediction models. Overall, the BTC/USDT trading pair offers an ideal combination of liquidity, data availability, and relative stability, making it an excellent choice for the investigation in this study.

This section details the process of data mining required to prepare candlestick data for modeling and predicting the returns of a cryptocurrency pair. The starting point is the generation and provision of candlestick data via the Binance API, specifically for the trading pair BTC/USDT (Bitcoin/US Dollar Tether). This API offers a comprehensive interface for querying historical market data, which includes essential information such as time, opening price, highest price, lowest price, closing price, and trading volume.

A Python function utilizing the Binance API is used to query the candlestick data. This function initializes a client for the Binance API using the provided API keys and then retrieves historical candlestick data for the BTC/USDT trading pair and the specified interval. The retrieved data is converted into a DataFrame, with only the first six columns being relevant: time, opening price, highest price, lowest price, closing price, and volume. The data is then formatted and converted into appropriate data types. This structured data forms the foundation for the subsequent steps of feature extraction and modeling.

An important aspect of processing this data is ensuring its completeness and correctness. In practice, certain candlesticks may be missing, leading to gaps in the data. These gaps can significantly affect the accuracy of the models. Therefore, missing candlesticks need to be identified and corrected. This can be achieved by filling in the missing timestamps with interpolated or constant values. A common approach to correction is to create a complete time index based on the specified interval and then fill in the missing timestamps in the DataFrame with the most recently available values.

By following these steps, it is ensured, that the candlestick data for BTC/USDT is complete and correct, providing a reliable foundation for the subsequent phases of feature engineering and modeling. Careful preparation and correction of the data are crucial to maximize the quality of predictions and ensure the performance of the models.

### B. Feature Engineering

In feature engineering, relevant features are extracted and processed from the candlestick data to improve the models' prediction accuracy. The essential steps of feature engineering in this work include data augmentation, prediction of returns, stationarity check, scaling, and the generation of sequences and labels.

### C. Data Augmentation

To enrich the information in the candlestick data, various technical indicators are calculated and integrated into the dataset. The key indicators include:

- **EMA12**: The Exponential Moving Average with a period of 12 days. The EMA smooths price data by giving more weight to recent data points, thus better identifying trends.
- **MACDline and MACDsignal**: The Moving Average Convergence Divergence (MACD) is a trend-following indicator. The MACDline is calculated as the difference between the 12-period EMA and the 26-period EMA. The MACDsignal is a 9-period EMA of the MACDline and is used to identify buy and sell signals.
- **%K and %D**: These values are part of the Stochastic Oscillator. %K measures the current price level relative to the price range over a specified period (typically 14), and %D is a moving average of %K.
- **RSI**: The Relative Strength Index measures the speed and change of price movements and is used to identify overbought or oversold conditions.
- **SMA12**: The Simple Moving Average over 12 periods, which is used to smooth price data and identify trends.

These indicators are calculated based on the closing prices of the candlestick data and added to the dataset to obtain a comprehensive feature spectrum.

*1) Returns:* The primary task is to predict the returns in future time steps $T$. The return is defined as the percentage change in the closing price between two consecutive time points:

$$\text{Return}_T = \frac{\text{Close}_T - \text{Close}_{T-1}}{\text{Close}_{T-1}} \tag{4}$$

This calculation quantifies the relative price change, which is essential for predicting future price movements. The neural network are fed with these returns to predict the return in $n$ future time steps. The input returns are based on hourly candlesticks, allowing for continuous and detailed analysis of price movements.

*2) Stationarity and Scaling:* For many time series models, it is important that the data is stationary, meaning their statistical properties, such as mean and variance, remain constant over time. To check the stationarity of the data, the Augmented Dickey-Fuller (ADF) test is used. The ADF test helps determine if a time series has a unit root, which is an indicator of non-stationarity. If the data are not stationary, they are transformed through differencing until stationarity is achieved. This transformation is necessary to ensure the stability of statistical properties.

Before scaling the data, it is important to ensure they are stationary. Scaling before checking for stationarity can result in mean and variance changes over time, distorting the scaling result. A MaxAbsScaler is used to scale the data to the range [-1, 1] without altering the original structure of the data. This is particularly useful for models, that are sensitive to the magnitude of the input data.

*3) Sequence and Label Generation:* The final step in feature engineering is generating sequences and corresponding labels. From the prepared data, we create fixed-length sequences that serve as input for the models. Each sequence consists of a set number of candlestick data points and the calculated technical indicators. The label for each sequence is the return in the subsequent time step $T$. This method ensures, that the models learn to predict future price movements based on a history of data.

When forming the labels, we ensure that the data at the output remains unscaled. This allows the neural network to fully utilize its regression properties and deliver more precise predictions. Unscaled labels ensure that the model learns and predicts the actual values directly, which is crucial for forecasting financial metrics such as returns. This approach improves the accuracy of the predictions and makes them easier to interpret.

Additionally, we implement the ability to adjust the length of the lookback sequences (the number of past time points considered for prediction) and the lookforward times (the number of future time steps to be predicted) individually. This allows us to find the optimal input sequence length and the appropriate prediction time for different models and datasets. This flexibility can further enhance prediction performance and better cater to the specific requirements of each use case.

These comprehensive steps in feature engineering ensure, that the data is optimally prepared and enriched, forming the foundation for subsequent modeling and prediction.

### D. Models

The basic building blocks of ANNs are neurons (also referred to as nodes, cells or units). A neuron is the representation of a function, that takes an input and computes a corresponding output. Inside of an ANN sets neurons of the same type are grouped into layers. There are three types of layers an ANN is built of

- **Input Layer**: Sets the number of input features
- **Hidden Layers**: An arbitrary number of different layers, that are not directly visible from the outside of the net
- **Output Layers**: Sets the number of output features

Each layer consists of a specified number of nodes and each node has a set of weighted inputs and an output [5].

Since the aim of this paper is to compare the performance of the different models, the theoretical principles of how they work, are only dealt with superficially. For more detailed explanations and information, please refer to the sources provided.

The models used in this paper, are a Multiple Layer Perceptron (MLP) ANN, a Convolutional Neural Network (CNN), a Recurrent Neural Network (RNN) and a special form of RNNs the Long Short-Term Memory (LSTM). The number of inputs and the hidden layers vary and are adapted to each model. We want to predict a single value, the price change of bitcoin in $T$ time steps, thus the number of outputs of all nets is one. Therefore last layer of all models consist of one node, that just linearly combines all outputs of the previous layer.

*1) MLP:* MLPs are the most commonly used ANN because of their simplicity. A MLP consists just of a set of Fully Connected Layers (FCLs), which is a layer, where each node of a layer takes every output of the previous layer as a weighted input. The output of the nodes of a FCL is recommended to be nonlinear [6]. The MLP model considered in this paper, is built of two fully connected layers, each consisting of 64 nodes. To make the MLP model capable of processing time series data, the feature sequences are flattened, that means, that the sequence of feature vectors are concatenated to one new vector of the size features times sequence length. Consequently the input size of the MLP is calculated dynamically dependent on the sequence length. The activation function for all nodes except the last one is the ReLU function.

*2) CNN:* A CNN is a ANN, that executes one or more convolution operations on the input data, in order to extract new features. Typically these kinds of nets are used for image processing, where a two dimensional convolution is performed. The main components of a CNN are the name giving convolutional and pooling layers. Convolutional layers perform a mathematical convolution by shifting a filter over the input data. Pooling layers devide the input data, here the input sequence into subsequences, where it chooses a single value for every feature in all subsequences and drops all other values [7]. In this case, the convolution is one dimensional and performed over the time axis, since the data is a time series. The CNN considered in this study consists of one single convolutional layer with a filter size of three, followed by a max pooling layer, that divides the sequence in sub sequences of the given size of also three and picks the highest value of every feature. The result of the pooling gets flattened and inserted into a FCL with 128 neurons, followed by one with 32 neurons. The activation function for all the nodes of the FCL is the ReLU function.

*3) RNN:* RNNs are specifically designed to process sequential data such as natural language or time series. Therefore recurrent neuron has a hidden state $h_t$. At each time step $t$ the recurrent node processes the input and its hidden state of the previous time step $h_{t-1}$ as input and updates the hidden state [8]. The RNN considered in this paper consists of one layer of 32 recurrent nodes, as well as one FCL of 64 nodes. The activation function of both the recurrent layer and the FCL is the ReLU function. The problem about simple RNNs is the so-called vanishing or exploding gradient [9]. This problem may occur when training over a large sequence of data for learning long time dependencies. The back propagation through time requires a repeated multiplication of the gradient by itself, which leads to the problem mentioned, if the gradient is very big or very small.

*4) LSTM:* A LSTM is a special type of RNN which faces the previously mentioned problems. The functionality and how these problems are avoided is described in [10].

The architecture of the LSTM used in this work is the same as the one of the RNN, with the only difference, that the recurrent layer is exchanged with a LSTM one.

### E. Training

With the term "training" of an ANN, the process of adjusting the parameters of the models (e.g. the different weights of a FCL) to fit the given problem. In general this is a classic optimization problem, therefore a loss function, which is to be optimized, needs to be formulated, an optimizing algorithm has to be chosen and a way of adjusting the parameters in the required way is needed.

The algorithm used for training is the so called "backpropagation". This is a efficient algorithm used to train ANNs for

supervised learning problems [11]. It consists of the following steps:

1) **Forward Pass**: The input data is passed through the network to create a prediction for the output.
2) **Computation of the Loss**: The quality of the prediction is quantified using a predefined loss function.
3) **Backward Pass**: The error is propagated backwards through the network to update the weights.

*1) Calculation of accuracy:* The models are trained to support decision making for buying or selling Bitcoin. For this decision a rough estimation of the price movement is considered as sufficient. More critical are predictions, which predict a change in the opposite direction than the actual price change. If there is an deviation between the predicted return and the actual return, but the direction of price change is correct, the prediction can still be assumed acceptable. In contrast, a prediction which indicates a price change in the wrong direction may be critical. To track the correct direction of the price change, we introduce a sign accuracy metric, which is shown in (5). *sign* is the signum function, that returns −1 for negative and +1 for positive values. *acc* is then the ratio of correct predicted price change direction to total number of predictions.

$$acc = \frac{1}{n} \cdot \sum_{i=1}^{n} \left| \frac{sign(y_i) + sign(\hat{y}_i)}{2} \right| \quad (5)$$

*2) Loss Function:* The purpose of the model is to predict the price change of the Bitcoin closing prices, which is a supervised learning regression task. For regression, ANNs typically a Mean Squared Error (MSE) is chosen as loss function. Which is, as the name proposes, the mean of the squared difference between the actual and the predicted value (see (6)). The total number of values is given by $n$, $y_i$ are the actual and $\hat{y}_i$ the corresponding predicted values.

$$mse = \frac{1}{n} \cdot \sum_{i=1}^{n} (y_i - \hat{y}_i)^2 \quad (6)$$

In order to achieve a high accuracy in price movement direction and absolute change prediction, we combine the MSE and the sign accuracy metric to get a suited loss function for the optimization, the *mse* is scaled by the inverse of the *acc*, in this way both metrics have a similar impact on the loss (see (7)). The parameter $\varepsilon$ is a small value (chosen as $10^{-6}$) added to the denominator to prevent division by zero and ensure numerical stability.

$$loss = \frac{mse}{acc + \varepsilon} \quad (7)$$

*3) Optimizer:* The optimizer chosen to minimize the loss function is the so-called "Adam" optimizer. It combines two extensions of the Stochastic Gradient Descend (SGD) and stands for Adaptive Moment Estimation [12].

*4) Parameters:* There are different parameters which may have an impact on the training result of a model. For most of these parameters, there are standard values, which are preset by the libraries offering tools for training, ANNs such as PyTorch in this case. Most of these parameters were not adjusted for this work because they have been preset values that have shown to be best practice. Nevertheless, there were two parameters, that were adjusted in a process of trial and error to gather the best performing outcomes.

An essential parameter for the optimizer is the so-called learning rate or step size. The optimizer determines a modified gradient of the loss function in dependency of the model parameters. This gradient is then multiplied by the negative value of the learning rate, to determine how each weight should be adjusted. A large learning rate may result in faster convergence, but may also lead to an overshoot, which causes a slower convergence or no convergence at all. As the name supposes, it is the base for the optimizer to calculate how "far" it goes in the direction it determined, that would minimize the loss function the most. The second parameter, that was adjusted, is the number of training epochs. In one epoch, the model is trained once on the entire training dataset. The results of an epoch are validated by evaluating the performance of the model on a validation dataset. One epoch means, that the model is trained once for the completed giving training dataset and evaluating the training results on a validation dataset. The number of epochs and the learning rate depend on each other, because a low learning rate leads to an expected lower convergence rate. This means, that the number of epochs should be increased and vice versa.

In a process of trial and error it was found, that a low learning rate of $10^{-5}$ combined with a number of epochs about $30$ produces the best performances.

IV. COMPARISON OF MODELS

*A. Used Data*

Initially, 5-minute candlesticks are used to make predictions. However, it becomes apparent, that networks work very slowly with this data because significantly more data have to be processed. Additionally, the results of the 5-minute candlesticks are significantly worse than those of the 1-hour candlesticks. This can be attributed to the fact, that the other features of the input tensor, consisting of technical indicators, are calculated on larger time units, which means, that these features do not contribute relevant information for prediction.

For this reason, a dataset with hourly candlesticks are selected for further analysis and predictions. This is shown in Figure 1, where the hourly BTCUSDT closing price data is depicted. A total of data from the last 400 days is included. The data split is done in a 70/20/10 ratio for training, validation, and test, as seen in Figure 1. The split is done sequentially, so that the majority of the oldest data is used for training, and the more recent data is used for validation and testing.

## B. Loss on Training and Validation Data

In Figure 2 the losses over the 30 training epochs for both the training and validation datasets is shown. The networks is trained here with a lookback of 24 hours for a 2-hour forecast. The Y-axis appears small because the network's prediction is a percentage return, which is scaled and displayed as a percentage.

Notably, despite a very small learning rate of $10^{-5}$, both the MLP and CNN show a very good fit to the validation data after the first epoch, compared to LSTM and RNN. This suggests overfitting, especially since the loss almost drops to zero in the further course. Even in the LSTM network, the loss quickly approaches zero in the later epochs, also indicating overfitting. The LSTM network's loss also quickly drops to zero in the subsequent epochs, which also indicates overfitting. Only the RNN network shows a continuous improvement over the training periods.

## C. Prediction, Mean Absolute Error, Prediction Accuracy

In Figure 3, the results of the predictions (in black) is compared to the actual returns (in red, labeled as "actual"). The networks are provided with a sequence length of 24 hours (historical data) and are tasked with predicting the return two hours into the future based on these 24-hour feature sequences.

Notably, the LSTM network exhibits small, almost negligible spikes in the training data prediction. In the test data, very few peaks are accurately predicted. From such a graph, it can be inferred, that the information density is not high enough for the network to make a clear decision, resulting in predictions that seem more random. In contrast, the RNN data shows better predictions in both the training and test data. Particularly in the test data, which represents the most recent data chronologically, it is evident that positive returns are better recognized by the network compared to negative returns.

*1) Mean Absolute Error:* In Figure 4, the Mean Absolute Errors (MAE) of the networks on the three datasets are shown, along with a comparison to the baseline, which simply uses the average absolute return of the time sequence as the prediction. A striking feature is the shape of the triplets for each network: there is a good fit on the training data, the test data are also well-fitted, but the MAE on the validation data is more than 50% higher than on the other two datasets. This can be attributed to the split of the dataset. Looking at the dataset split in Figure 1, a pattern can be observed. The training and test data exhibit a ramp pattern, meaning we start with a small value in the dataset and end with a higher one, so the returns are generally positive on average. The validation dataset can be described by a hill pattern: it rises at the beginning and then falls again, so the values are back to the same magnitude as at the start, meaning the average return is more or less zero. This indicates that the network, despite the measures taken for stationarity, still perceives and tries to follow trends.

Overall, in Figure 4, the four tested networks are compared to the baseline, which is calculated as the average absolute return of the specified input sequence length. In this case, the baseline is calculated from the last 24 hours. For example, an absolute error of 0.6% is calculated for the return of the baseline on the training data with an input sequence of 24 hours. For the other networks, the reported error calculated for the individual dataset provides information about the prediction accuracy. For the other networks, the given error calculated for the individual dataset provides information on the inaccuracy of the prediction. This means, that, if a return to be predicted (label) is smaller than the MAE, no accurate statement can be made about whether a positive or negative return can be expected.

In addition to the aforementioned characteristic of the prediction thresholds, it can be observed from Figure 4 that the MAEs of the networks are significantly below that of the baseline. Here, the MAE values are on average 30% smaller compared to the baseline. The LSTM and RNN perform best on the test dataset with a 0.41% MAE with the chosen input sequence of 24 hours and a prediction of two hours. The figure clearly demonstrates that prediction with the networks is feasible.

*2) Prediction Accuracy:* Using the previously explained threshold criterion, which utilizes the MAE (Mean Absolute Error), we now investigate whether different sequence lengths have an impact on the accuracy of return predictions. The accuracy in Table II is based on the MAE from Table I, which serves as the threshold, excluding all returns smaller than this threshold. This exclusion occurs because the inaccuracy in small returns could more easily lead to erroneous return predictions. In the second instance, we check which of the returns exceeding the threshold were correctly predicted by the respective network. The percentage of correctly predicted returns greater than the threshold represents the accuracy. In this part, four sequence lengths are compared: 168 hours (7 days), 72 hours (3 days), 24 hours, and 12 hours, with a return prediction of 2 hours for all.

The depiction in Figure 3 shows most predictions made by the nets are relatively small and there are only few spikes. Those predicted spikes on the other hand seem to be relatively accurate. Because of that, it is assumed, that the small predictions can be interpreted as noise ad should not be considered for decision-making. Instead, a new metric, that calculates the accuracy of all predictions whose values are above a threshold. Although this reduces the number of possible indications, in exchange the quality of the predictions should be increased.

This hypothesis then was checked on the models for different lookback sequences and a prediction on the price change in two hours. The threshold was chosen to be the MAE listed in Table I, all predictions whose absolute value is lower than the

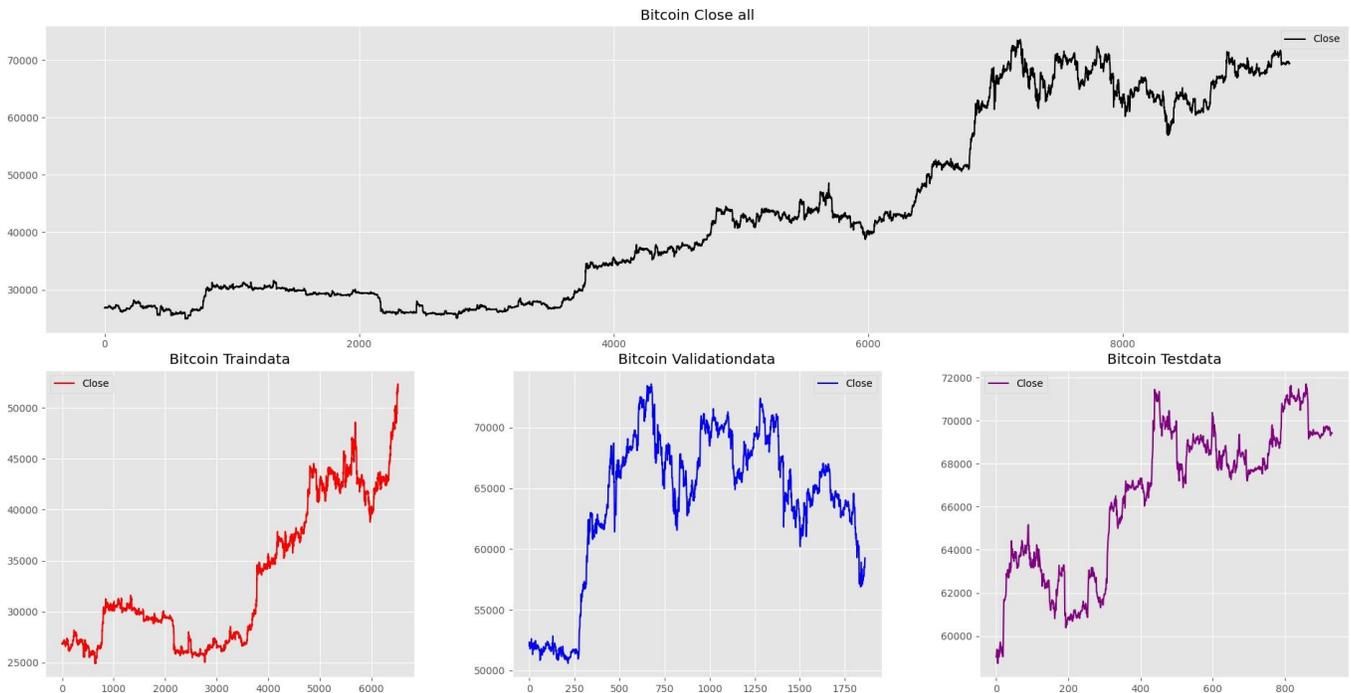

Fig. 1: Close Data of BTC USDT for the last 400 Days and the Split

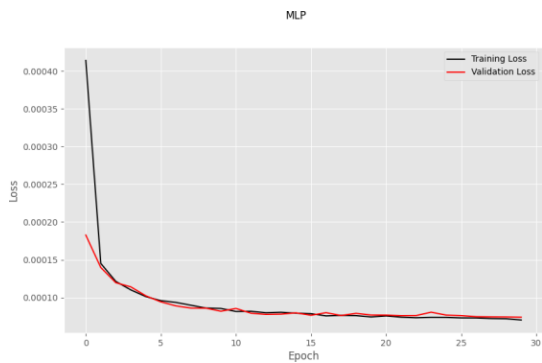

(a) Loss over epochs of the MLP

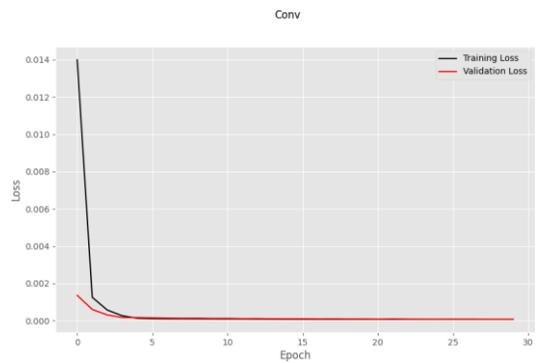

(b) Loss over epochs of the CNN

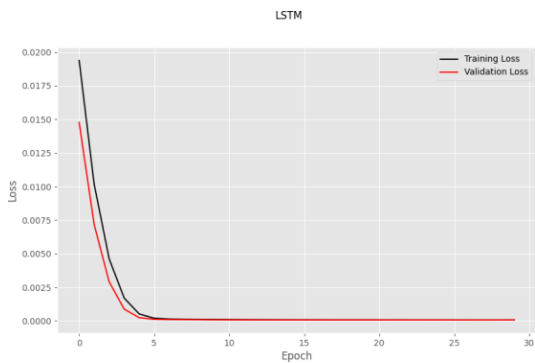

(c) Loss over epochs of the LSTM

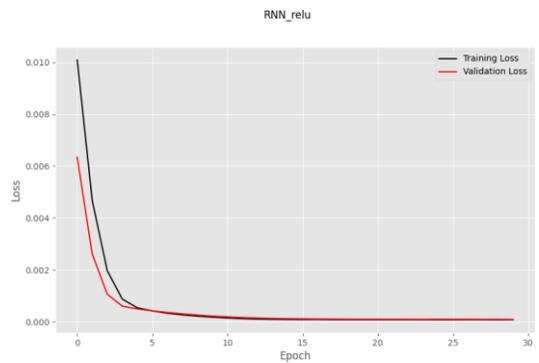

(d) Loss over epochs of the RNN

Fig. 2: Loss of Train and Validation data over the different Trainingsepochs

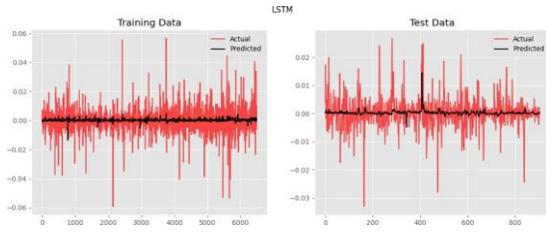

(a) Prediction of LSTM and actual data of 2 hours forecast

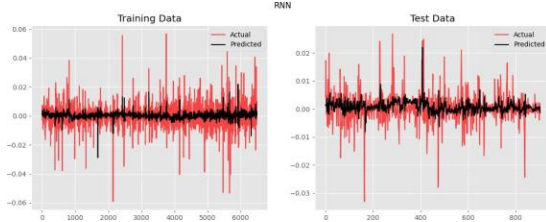

(b) Prediction of RNN and actual data of 2 hours forecast

Fig. 3: Two Hours Forecast of RNN and LSTM

|  |  | 168h [%] | 72h [%] | 24h [%] | 12h [%] |
|---|---|---|---|---|---|
| MLP | train | 0.35 | 0.38 | 0.39 | 0.39 |
|  | val | 0.72 | 0.7 | 0.69 | 0.69 |
|  | test | 0.45 | 0.44 | 0.62 | 0.44 |
| RNN | train | 0.37 | 0.38 | 0.38 | 0.38 |
|  | val | 0.65 | 0.63 | 0.64 | 0.64 |
|  | test | 0.39 | 0.4 | 0.42 | 0.41 |
| LSTM | train | 0.37 | 0.37 | 0.4 | 0.37 |
|  | val | 0.63 | 0.62 | 0.64 | 0.62 |
|  | test | 0.38 | 0.38 | 0.41 | 0.41 |
| Conv | train | 0.37 | 0.36 | 0.41 | 0.38 |
|  | val | 0.67 | 0.67 | 0.71 | 0.65 |
|  | test | 0.4 | 0.43 | 0.49 | 0.42 |

TABLE I: Mean absolute error of the predictions for 2h price change in %

|  |  | 168h [%] | 72h [%] | 24h [%] | 12h [%] |
|---|---|---|---|---|---|
| MLP | train | 80 | 62.1 | 60 | 53.3 |
|  | val | 53.2 | 53.2 | 49.3 | 51.6 |
|  | test | 54.8 | 60 | 63.3 | 52.9 |
| RNN | train | 55.3 | 63.5 | 47.6 | 49.5 |
|  | val | 41.9 | 62.5 | 50 | 48.3 |
|  | test | 47.8 | 45.5 | 42.0 | 53.3 |
| LSTM | train | 52.9 | 20 | 44.9 | 45 |
|  | val | 100 | 80 | 28.2 | 72.7 |
|  | test | 0 | 0 | 33.3 | 50 |
| Conv | train | 76.3 | 75.9 | 53.4 | 60.8 |
|  | val | 58.5 | 45.7 | 46.3 | 52.5 |
|  | test | 52.5 | 52.1 | 48.8 | 37.2 |

TABLE II: Accuracy of predicted moving direction on predictions greater than the mean absolute error in %

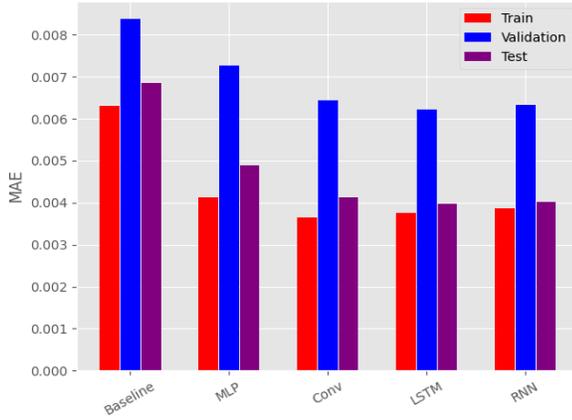

Fig. 4: Mean-Absolut-Error between the Baseline and different NNs for a 24h Lookbacksequenz und 2h forecast

corresponding MAE are neglected. On this reduced prediction sets, the accuracy is calculated again. The results are shown in Table II.

The MAE on the validation dataset is significantly larger than the ones on the train and test dataset for almost all models on all sequence length. This can be attributed to the similarity of the training and test data mentioned before.

It can be seen, that the accuracy of the threshold predictions is at most on two of the three datasets significantly larger than 50% which is necessary for a model to qualify as a reliable indicator. The outliers in accuracy of 0 or 100% of the LSTM can be explained by the low number of predictions of the model that exceed the threshold. If there is for example only one prediction that satisfies the condition, the accuracy is either 0 or 100%. The high accuracy, which can be partially seen for the train dataset, can be the effects of overfitting.

The overall best performing configurations are the MLP for the forecasting periods of 168h, 72h and 12h as well as the CNN for 168h, since all accuracy are above the required minimum of 50%. Nevertheless, there is no model that can predict the direction of price change with an accuracy significantly above 50% on all three datasets; therefore the previously formulated hypothesis is not fulfilled.

## V. CONCLUSION

The investigation into the prediction accuracy of various sequence lengths for forecasting cryptocurrency returns shows that none of the tested model structures delivered satisfactory prediction performance. Despite the application of the MAE threshold criterion, which aims to improve the accuracy of the predictions by excluding small, error-prone returns, the results remained unsatisfactory.

It is found that the prediction accuracy of most models is only slightly above random chance. This suggests that the models have difficulties in recognizing and predicting reliable trends in the data. In particular, the limited number of features, which are exclusively extracted from candlestick data, seem to be a significant limiting factor. These features may not capture the complexity and deeper patterns of the market necessary for more accurate predictions.

Additionally, the analysis shows signs of overfitting, particularly with the LSTM models, indicating that the models might have learned the training data too well without capturing generalizable patterns that could be applied to new, unseen data. This problem might be alleviated by implementing regularization techniques such as L1/L2 regularization or dropout, but the current results suggest that this alone is insufficient.

Another issue is data leakage, which can compromise the validity of the model results. Careful review and cleansing of data sources are necessary to ensure that no future information leaks into the training data, artificially boosting model performance.

In conclusion, predictions based solely on candlestick data are not optimal. To improve prediction accuracy, additional data sources and features should be included, such as number of trades, technical indicators, and fundamental data. Furthermore, different network architectures such as transformer models or hybrid approaches combining various neural networks could be tested to enhance prediction performance.

Future work should therefore focus on integrating a broader range of input data and employing more advanced modeling approaches to better capture the complexity of financial markets and enable more accurate predictions.